FLORENTIN SMARANDACHE

# α-Discounting Method for Multi-Criteria Decision Making (α-D MCDM)

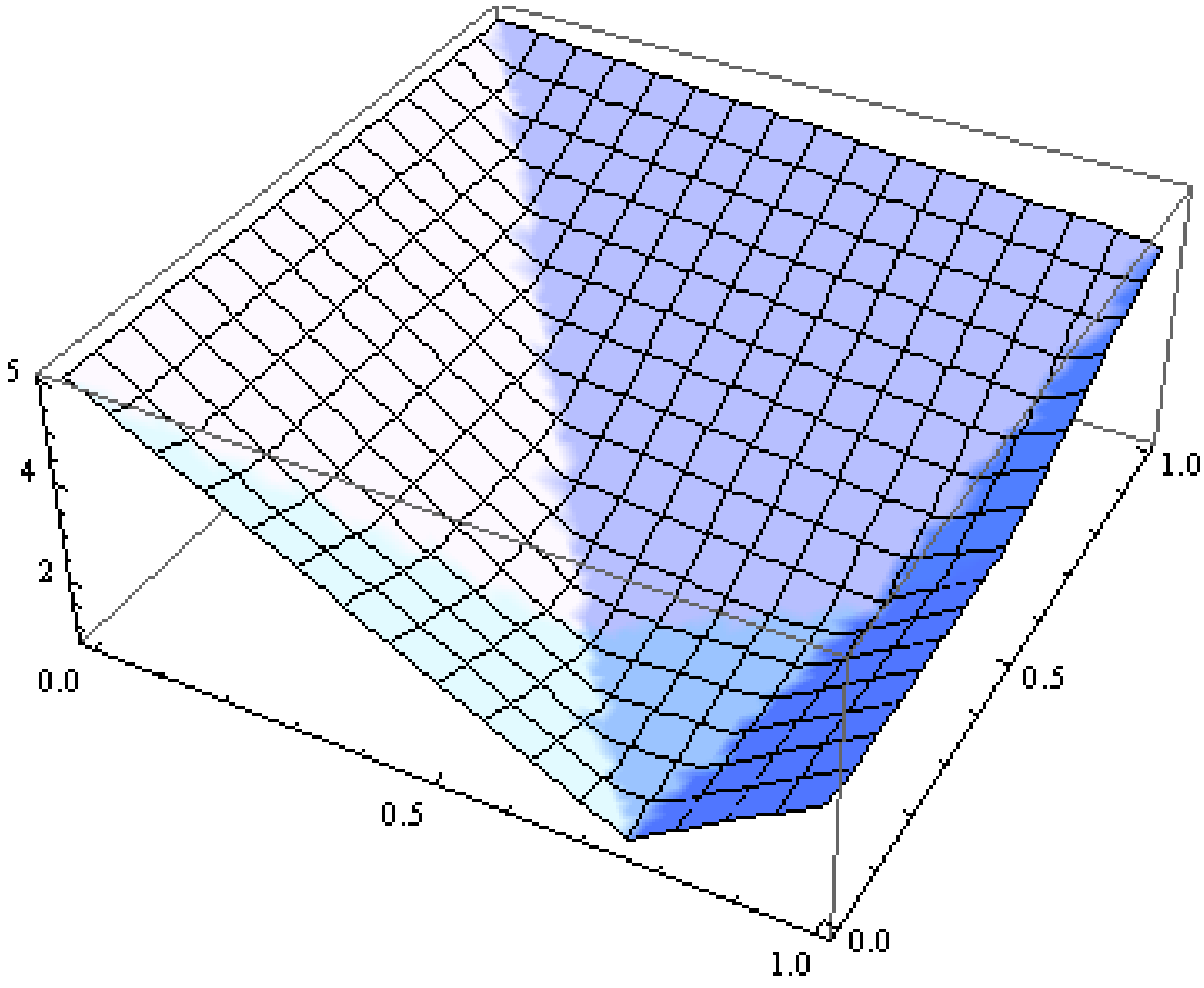

2015

FLORENTIN SMARANDACHE

# α-Discounting Method for Multi-Criteria Decision Making (α-D MCDM)

*The ideas in this book were previously published or presented at conferences, as it follows:*

1. *α-Discounting Method for Multi-Criteria Decision Making (α-D MCDM)*, by Florentin Smarandache, Fusion 2010 International Conference, Edinburgh, Scotland, 26-29 July, 2010; published in "Review of the Air Force Academy / The Scientific Informative Review", No. 2, 29-42, 2010; also presented at Osaka University, Department of Engineering Science, Inuiguchi Laboratory, Japan, 10 January 2014.
2. *Three Non-linear α-Discounting MCDM-Method Examples*, Proceedings of The 2013 International Conference on Advanced Mechatronic Systems (ICAMechS 2013), Luoyang, China, September 25-27, pp. 174-176, 2013; and in Critical Review, Center for Mathematics of Uncertainty, Creighton University, USA, Vol. VIII, 32-36, 2014.

# Table of Content

# Preface

In this book we introduce a new procedure called α-Discounting Method for Multi-Criteria Decision Making (α-D MCDM), which is as an alternative and extension of Saaty's Analytical Hierarchy Process (AHP). It works for any number of preferences that can be transformed into a system of homogeneous linear equations. A degree of consistency (and implicitly a degree of inconsistency) of a decision-making problem are defined. α-D MCDM is afterwards generalized to a set of preferences that can be transformed into a system of linear and/or non-linear homogeneous and/or non-homogeneous equations and/or inequalities.

Many consistent, weak inconsistent, and strong inconsistent examples are given.

In Chapter 1, it is presented the general idea of α-D MCDM, which is to assign non-null positive parameters α1, α2, ..., αp to the coefficients in the right-hand side of each preference that diminish or increase them in order to transform the above linear homogeneous system of equations which has only the null-solution, into a system having a particular non-null solution.

After finding the general solution of this system, the principles used to assign particular values to all parameters α's is the second important part of α-D, yet to be deeper investigated in the future.

In the current chapter we herein propose the Fairness Principle, i.e. each coefficient should be discounted with the same percentage (we think this is fair: not making any favouritism or unfairness to any coefficient), but the reader can propose other principles.

For consistent decision-making problems with pairwise comparisons, α-Discounting Method together with the Fairness Principle give the same result as AHP.

But for weak inconsistent decision-making problem, α - Discounting together with the Fairness Principle give a different result from AHP.

α-Discounting/Fairness-Principle together give a justifiable result for strong inconsistent decision-making problems with two preferences and two criteria; but for more than two preferences with more than two criteria and the Fairness Principle has to be replaced by another principle of assigning numerical values to all parameters α's.

In Chapter 2 we present three new examples of using the α-Discounting Multi-Criteria Decision Making Method in solving non-linear problems involving algebraic equations and inequalities in the decision process.

Chapter 3 is an extension of our previous work on α-Discounting Method for MCDM from crisp numbers to intervals.

*The author.*

# α-Discounting Method for Multi-Criteria Decision Making (α-D MCDM)

## Abstract

In this chapter we introduce a new procedure called α-Discounting Method for Multi-Criteria Decision Making (α-D MCDM), which is as an alternative and extension of Saaty's Analytical Hierarchy Process (AHP). It works for any number of preferences that can be transformed into a system of homogeneous linear equations. A degree of consistency (and implicitly a degree of inconsistency) of a decision-making problem are defined.

α-D MCDM is generalized to a set of preferences that can be transformed into a system of linear and/or non-linear homogeneous and/or non-homogeneous equations and/or inequalities.

Many consistent, weak inconsistent, and strong inconsistent examples are given.

## Introduction

α-Discounting Method for Multi-Criteria Decision Making (α-D MCDM) is an alternative and extension of Saaty's Analytical Hierarchy Process (AHP) - see [1-11] for more information on AHP since it is not the main subject of this chapter. It works not only for preferences that are pairwise comparisons of criteria as AHP does, but for preferences of any n-wise (with n≥2) comparisons of criteria that can be expressed as linear homogeneous equations.

The general idea of α-D MCDM is to assign non-null positive parameters $\alpha_1, \alpha_2, \ldots, \alpha_p$ to the coefficients in the right-hand side of each preference that diminish or increase them in order to transform the above linear homogeneous system of equations which has only the null-solution, into a system having a particular non-null solution.

After finding the general solution of this system, the principles used to assign particular values to all parameters α's is the second important part of α-D, yet to be deeper investigated in the future. In the current chapter we herein propose the Fairness Principle, i.e. each coefficient should be discounted with the same percentage (we think this is fair: not making any favouritism or unfairness to any coefficient), but the reader can propose other principles.

For consistent decision-making problems with pairwise comparisons, α-Discounting Method together with the Fairness Principle give the same result as AHP. But for weak inconsistent decision-making problem, α-Discounting together with the Fairness Principle give a different result from AHP.

α-Discounting/Fairness-Principle together give a justifiable result for strong inconsistent decision-making problems with two preferences and two criteria; but for more than two preferences with more than two criteria and the Fairness Principle has to be replaced by another principle of assigning numerical values to all parameters α's.

Since Saaty's AHP is not the topic of this chapter, we only recall the main steps of applying this method, so the results of α-D MCDM and of AHP could be compared.

AHP works only for pairwise comparisons of criteria, from which a square Preference Matrix, A (of size $n \times n$), is built. Then one computes the maximum eigenvalue $\lambda_{max}$ of A and its corresponding eigenvector.

If $\lambda_{max}$ is equal to the size of the square matrix, then the decision-making problem is consistent, and its corresponding normalized eigenvector (Perron-Frobenius vector) is the priority vector.

If $\lambda_{max}$ is strictly greater than the size of the square matrix, then the decision-making problem is inconsistent. One raise to the second power matrix A, and again the resulted matrix is raised to the second power, etc. obtaining the sequence of matrices $A^2$, $A^4$, $A^8$, …, etc. In each case, one computes the maximum eigenvalue and its associated normalized eigenvector, until the difference between two successive normalized eigenvectors is smaller than a given threshold. The last such normalized eigenvector will be the priority vector.

Saaty defined the Consistency Index as:

$$CI(A) = \frac{\lambda_{max}(A) - n}{n - 1},$$

where n is the size of the square matrix A.

# α-Discounting Method for Multi-Criteria Decision Making (α-D MCDM)

## Description of α-D MCDM

The general idea of this chapter is to discount the coefficients of an inconsistent problem to some percentages in order to transform it into a consistent problem.

Let the Set of Criteria be $C = \{C_1, C_2, \ldots, C_n\}$, with $n \geq 2$, and the Set of Preferences be $P = \{P_1, P_2, \ldots, P_m\}$, with $m \geq 1$.

Each preference $P_i$ is a linear homogeneous equation of the above criteria $C_1, C_2, \ldots, C_n$:

$$P_i = f(C_1, C_2, \ldots, C_n).$$

We need to construct a basic belief assignment (bba):

$$m: C \to [0, 1]$$

such that $m(C_i) = x_i$ , with $0 \leq x_i \leq 1$, and

$$\sum_{i=1}^{n} m(C_i) = \sum_{i=1}^{n} x_i = 1.$$

We need to find all variables $x_i$ in accordance with the set of preferences P.

Thus, we get an $m \times n$ linear homogeneous system of equations whose associated matrix is

$$A = (a_{ij}),\ 1 \leq i \leq m \text{ and } 1 \leq j \leq n.$$

In order for this system to have non-null solutions, the rank of the matrix A should be strictly less than n.

## Classification of Linear Decision-Making Problems

a) We say that a **linear decision-making problem is consistent** if, by any substitution of a variable $x_i$ from an equation into another equation, we get a result in agreement with all equations.

b) We say that a **linear decision-making problem is weakly inconsistent** if by at least one substitution of a variable $x_i$ from an equation into another equation we get a result in disagreement with at least another equation in the following ways:

$$\text{(WD1)} \quad \begin{Bmatrix} x_i = k_1 \cdot x_j, k > 1; \\ x_i = k_2 \cdot x_j, k_2 > 1, k_2 \neq k_1 \end{Bmatrix}$$

or

$$\text{(WD2)} \quad \begin{Bmatrix} x_i = k_1 \cdot x_j, 0 < k < 1; \\ x_i = k_2 \cdot x_j, 0 < k_2 < 1, k_2 \neq k_1 \end{Bmatrix}$$

or

$$\text{(WD3)} \quad \{x_i = k \cdot x_i, k \neq 1\}$$

(WD1)-(WD3) are weak disagreements, in the sense that for example a variable x > y always, but with different ratios (for example: x=3y and x=5y).

c) All disagreements in this case should be like (WD1)-(WD3).

We say that a **linear decision-making problem is strongly inconsistent** if, by at least one substitution of a variable $x_i$ from an equation into another equation, we get a result in disagreement with at least another equation in the following way:

$$\text{(SD4)} \quad \begin{Bmatrix} x_i = k_1 \cdot x_j; \\ x_i = k_2 \cdot x_j, \end{Bmatrix}$$

with $0 < k_1 < 1 < k_2$ or $0 < k_2 < 1 < k_1$ (i.e. from one equation one gets $x_i < x_j$ while from the other equation one gets the opposite inequality: $x_j < x_i$).

At least one inconsistency like (SD4) should exist, no matter if other types of inconsistencies like (WD1)-(WD3) may occur or not.

## Compute the determinant of A

a) If det(A)=0, the decision problem is consistent, since the system of equations is dependent.

It is not necessarily to parameterize the system. {In the case we have parameterized, we can use the Fairness Principle – i.e. setting all parameters equal $\alpha_1 = \alpha_2 = \ldots = \alpha_p = \alpha > 0$}.

Solve this system; find its general solution.

Replace the parameters and secondary variables, getting a particular solution.

Normalize this particular solution (dividing each component by the sum of all components).

Wet get the priority vector (whose sum of its components should be 1).

b) If det(A) $\neq$ 0, the decision problem is inconsistent, since the homogeneous linear system has only the null-solution.

b1) If the inconsistency is weak, then parameterize the right-hand side coefficients, and denote the system matrix A(α).

Compute det(A(α)) = 0 in order to get the parametric equation.

If the Fairness Principle is used, set all parameters equal, and solve for $\alpha > 0$.

Replace $\alpha$ in A(α) and solve the resulting dependent homogeneous linear system.

Similarly as in a), replace each secondary variable by 1, and normalize the particular solution in order to get the priority vector.

b2) If the inconsistency is strong, the Fairness Principle may not work properly. Another approachable principle might be designed.

Or, get more information and revise the strong inconsistencies of the decision-making problem.

## Comparison between AHP and α-D MCDM

a) α-D MCDM's general solution includes all particular solutions, that of AHP as well;

b) α-D MCDM uses all kind of comparisons between criteria, not only paiwise comparisons;

c) for consistent problems, AHP and α-D MCDM/Fairness-Principle give the same result;

d) for large inputs, in α-D MCDM we can put the equations under the form of a matrix (depending on some parameters alphas), and then compute the determinant of the matrix which should be zero; after that, solve the system (all can be done on computer using math software); the software such as MATHEMATICA and MAPPLE for example can do these determinants and calculate the solutions of this linear system;

e) α-D MCDM can work for larger classes of preferences, i.e. preferences that can be transformed in homogeneous linear equations, or in non-linear equations and/or inequalities – see more below.

## Generalization of α-D MCDM

Let each preference be expressed as a linear or non-linear equation or inequality. All preferences together will form a system of linear/non-linear equations/inequalities, or a mixed system of equations and inequalities.

Solve this system, looking for a strictly positive solution (i.e. all unknowns $x_i > 0$). Then normalize the solution vector.

If there are more such numerical solutions, do a discussion: analyze the normalized solution vector in each case.

If there is a general solution, extract the best particular solution by replacing the secondary variables by some numbers such that the resulting particular solution is positive, and then normalizing.

If there is no strictly positive solution, parameterize the coefficients of the system, find the parametric equation, and look for some principle to apply in order to find the numerical values of the parameters α's

A discussion might also be involved. We may get undetermined solutions.

## Degrees of Consistency and Inconsistency in α-D MCDM/Fairness-Principle

For $\alpha$-D MCDM/Fairness-Principle in consistent and weak consistent decision-making problems, we have the followings:

a) If $0 < \alpha < 1$, then $\alpha$ is the **degree of consistency** of the decision-making problem, and $\beta = 1-\alpha$ is the **degree of inconsistency** of the decision-making problem.

b) If $\alpha > 1$, then $1/\alpha$ is the **degree of consistency** of the decision-making problem, and $\beta = 1-1/\alpha$ is the **degree of inconsistency** of the decision-making problem.

## Principles of α-D MCDM (Second Part)

a) In applications, for the second part of $\alpha$-D Method, the Fairness Principle can be replaced by other principles.

*Expert's Opinion*. For example, if we have information that a preference's coefficient should be discounted twice more than another coefficient (due to an expert's opinion), and another preference's coefficient should be discounted a third of another one, then appropriately we set for example: $\alpha_1 = 2\alpha_2$ and respectively $\alpha_3 = (1/3)\alpha_4$, etc. in the parametric equation.

b) For $\alpha$-D/Fairness-Principle or Expert's Opinion.

Another idea herein is to set a **threshold of consistency** $t_c$ (or implicitly **a threshold of inconsistency** $t_i$). Then, if the degree of consistency is smaller than a required $t_c$, the Fairness Principle or Expert's Opinion (whichever was used) should be discharged, and another principle of finding all parameters $\alpha$'s should be designed; and similarly if the degree of inconsistency is bigger than $t_i$.

c) One may measure the <u>system's accuracy</u> (or <u>error</u>) for the case when all m preferences can be transformed into equations; for example, preference $P_i$ is transformed into an equation $f_i(x_1, x_2, ..., x_n)=0$; then we need to find the unknowns $x_1, x_2, ..., x_n$ such that:

$$e(x_1, x_2, ..., x_n) = \sum_{i=1}^{m} |f_i(x_1, x_2, ..., x_n)| \text{ is minimum,}$$

where “e(…)” means error.

Calculus theory (partial derivatives) can be used to find the minimum (if this does exist) of a function of n variables, $e(x_1, x_2, \ldots, x_n)$, with $e: R_+^n \rightarrow R_+$.

For consistent decision-making problems the system’s accuracy/error is zero, so we get the exact result.

We prove this through the fact that the normalized priority vector $[a_1\ a_2\ \ldots\ a_n]$, where $x_i=a_i > 0$ for all i, is a particular solution of the system $f_i(x_1, x_2, \ldots, x_n)=0$ for i=1, 2, …, m; therefore:

$$\sum_{i=1}^{m} |f_i(a_1, a_2, \ldots, a_n)| = \sum_{i=1}^{m} |0| = 0.$$

But, for inconsistent decision-making problems we find approximations for the variables.

## Extension of α-D MCDM (Non-Linear α-D MCDM)

It is not difficult to generalize the α-D MCDM for the case when the preferences are non-linear homogeneous (or even non-homogeneous) equations.

This non-linear system of preferences has to be dependent (meaning that its general solution – its main variables - should depend upon at least one secondary variable).

If the system is not dependent, we can parameterize it in the same way. Then, again, in the second part of this Non-Linear α-D MCDM we assign some values to each of the secondary variables (depending on extra-information we might receive), and we also need to design a principle which will help us to find the numerical values for all parameters.

We get a particular solution (such extracted from the general solution), which normalized will produce our priority vector.

Yet, the Non-Linear α-D MCDM is more complicated, and depends on each non-linear decision-making problem.

Let us see some examples.

## Consistent Example 1

A. *We use the α-D MCDM*. Let the Set of Preferences be:

$$\{C1, C2, C3\}$$

and The Set of Criteria be:

1. $C1$ is 4 times as important as $C2$.
2. $C2$ is 3 times as important as $C3$.
3. $C3$ is one twelfth as important as $C1$.

Let $m(C1) = x$, $m(C2) = y$, $m(C3) = z$.

The linear homogeneous system associated to this decision-making problem is:

$$\begin{cases} x = 4y \\ y = 3z \\ z = \dfrac{x}{12} \end{cases}$$

whose associated matrix $A_1$ is:

$$\begin{pmatrix} 1 & -4 & 0 \\ 0 & 1 & -3 \\ -1/12 & 0 & 1 \end{pmatrix},$$

whence $\det(A_1) = 0$, so the DM problem is consistent.

Solving this homogeneous linear system we get its general solution that we set as a solution vector [12z 3z z], where z can be any real number (z is considered a secondary variable, while x=12z and y=3z are main variables).

Replacing z=1 into the solution vector, the solution vector becomes [12 3 1], and then normalizing (dividing by 12+3+1=16 each vector component) we get the priority vector: [12/16 3/16 1/16], so the preference will be on C1.

B. *Using AHP, we get the same result.*

The preference matrix is:

$$\begin{pmatrix} 1 & 4 & 12 \\ 1/4 & 1 & 3 \\ 1/12 & 1/3 & 1 \end{pmatrix}$$

whose maximum eigenvalue is $\lambda_{\max} = 3$ and its corresponding normalized eigenvector (Perron-Frobenius vector) is [12/16 3/16 1/16].

C. *Using Mathematica 7.0 Software*:

Using MATHEMATICA 7.0 software, we graph the function:

$h(x,y) = |x-4y|+|3x+4y-3|+|13x+12y-12|$, with $x,y \in [0,1]$, which represents the consistent decision-making problem's associated system:

$x/y=4$, $y/z=3$, $x/z=12$, and $x+y+z=1$, $x>0$, $y>0$, $z>0$.

In[1]:=

Plot3D[Abs[x-4y]+Abs[3x+4y-3]+Abs[13x+12y-12],{x,0,1},{y,0,1}]

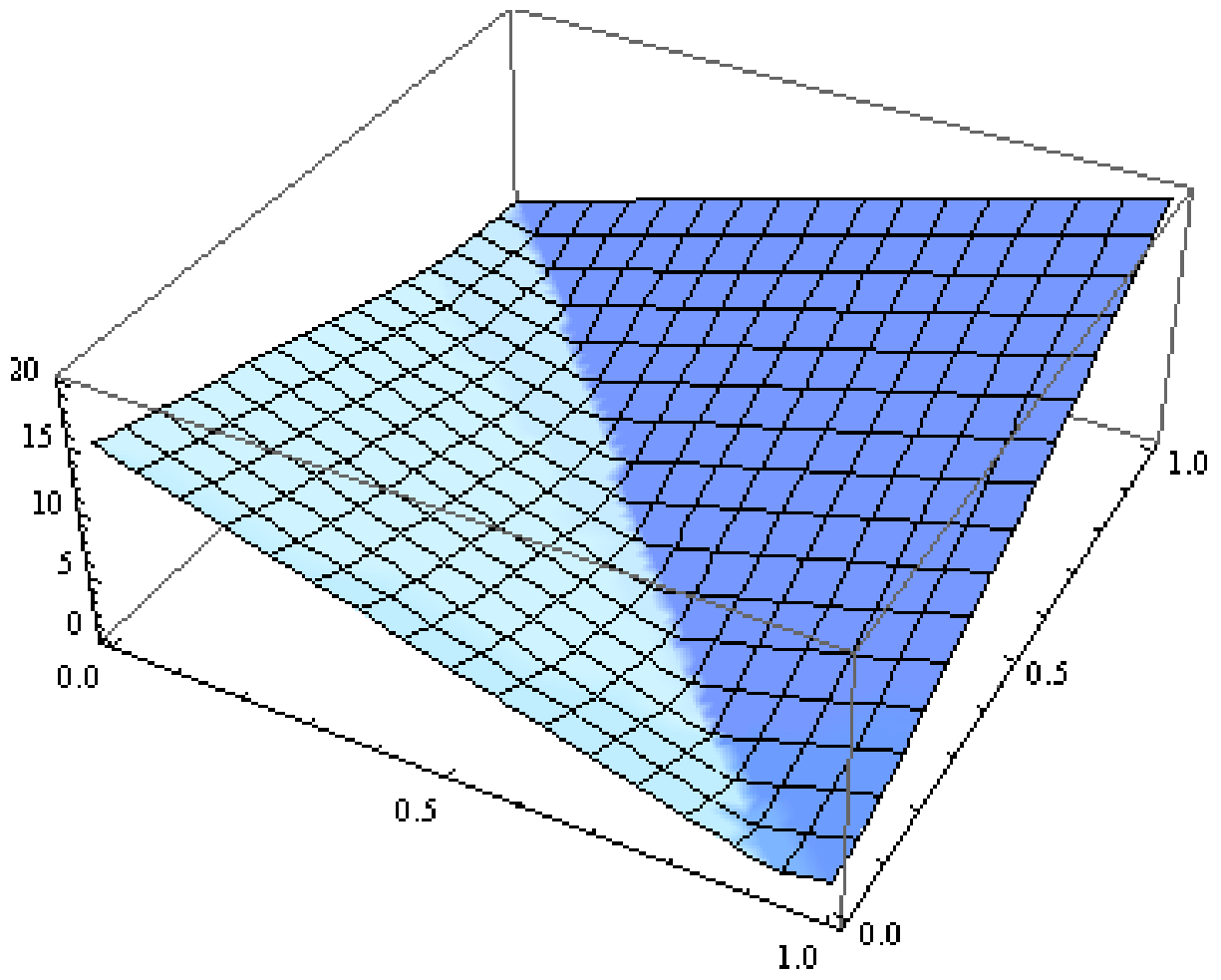

The minimum of this function is zero, and occurs for x=12/16, y=3/16.

If we consider the original function of three variables associated with h(x,y) we have:

H(x,y, z) = |x-4y|+|y-3z|+|x-12z|, x+y+z=1, with x,y,z $\in$ [0,1], we also get the minimum of H(x,y,z) being zero, which occurs for x=12/16, y=3/16, z=1/16.

## Weak Inconsistent Examples where AHP does not work

The Set of Preferences is $\{C1, C2, C3\}$.

### Weak Inconsistent Example 2

*α-D MCDM method.*

The Set of Criteria is:

1. $C1$ is as important as 2 times $C2$ plus 3 times $C3$.
2. $C2$ is half as important as $C1$.
3. $C3$ is one third as important as $C1$.

Let $m(C1) = x$, $m(C2) = y$, $m(C3) = z$;

$$\begin{cases} x = 2y + 3z \\ y = \dfrac{x}{2} \\ z = \dfrac{x}{3} \end{cases}$$

AHP cannot be applied on this example because of the form of the first preference, which is not a pairwise comparison.

If we solve this homogeneous linear system of equations as it is, we get x=y=z=0, since its associated matrix is:

$$\begin{pmatrix} 1 & -2 & -3 \\ -1/2 & 1 & 0 \\ -1/3 & 0 & 1 \end{pmatrix} = -1 \neq 0$$

but the null solution is not acceptable since the sum x+y+z has to be 1.

Let us parameterise each right-hand side coefficient and get the general solution of the above system:

$$\begin{cases} x = 2\alpha_1 y + 3\alpha_2 z & (1) \\ y = \dfrac{\alpha_3}{2} x & (2) \\ z = \dfrac{\alpha_4}{3} x & (3) \end{cases}$$

where $\alpha_1, \alpha_2, \alpha_3, \alpha_4 > 0$.

Replacing (2) and (3) in (1) we get

$$x = 2\alpha_1 \left( \frac{\alpha_3}{2} x \right) + 3\alpha_2 \left( \frac{\alpha_4}{3} x \right)$$

$$1 \cdot x = (\alpha_1\alpha_3 + \alpha_2\alpha_4) \cdot x$$

whence

$\alpha_1\alpha_3 + \alpha_2\alpha_4 = 1$ (parametric equation) (4)

The general solution of the system is:

$$\begin{cases} y = \dfrac{\alpha_3}{2} x \\ z = \dfrac{\alpha_4}{3} x \end{cases}$$

whence the priority vector:

$$\begin{bmatrix} x & \dfrac{\alpha_3}{2} x & \dfrac{\alpha_4}{3} x \end{bmatrix} \to \begin{bmatrix} 1 & \dfrac{\alpha_3}{2} & \dfrac{\alpha_4}{3} \end{bmatrix}.$$

Fairness Principle: discount all coefficients with the same percentage: so, replace

$\alpha_1 = \alpha_2 = \alpha_3 = \alpha_4 = \alpha > 0$ in (4) we get $\alpha^2 + \alpha^2 = 1$,

whence $\alpha = \dfrac{\sqrt{2}}{2}$.

Priority vector becomes:

$$\begin{bmatrix} 1 & \dfrac{\sqrt{2}}{4} & \dfrac{\sqrt{2}}{6} \end{bmatrix}$$

and normalizing it:

$$\begin{bmatrix} 0.62923 & 0.22246 & 0.14831 \end{bmatrix}$$
$$\begin{matrix} C1 & C2 & C3 \\ x & y & z \end{matrix}$$

Preference will be on C1, the largest vector component.

Let us verify it:

$\dfrac{y}{x} \cong 0.35354$ instead of 0.50, i.e. $\dfrac{\sqrt{2}}{2} = 70.71\%$ of the original.

$\frac{z}{x} \cong 0.23570$ instead of 0.33333, i.e. 70.71% of the original.

$x \cong 1.41421y + 2.12132z$ instead of $2y + 3z$, i.e. 70.71% of 2 respectively 70.71% of 3.

So, it was a fair discount for each coefficient.

*Using Mathematica 7.0 Software.*

Using MATHEMTICA 7.0 software, we graph the function:

g(x,y) = |4x-y-3|+|x-2y|+|4x+3y-3|, with x,y $\in$ [0,1],

which represents the weak inconsistent decision-making problem's associated system:

x-2y-3z=0, x-2y=0, x-3z=0, and x+y+z=1, x>0, y>0, z>0.

by solving z=1-x-y and replacing it in

G(x,y,z)= |x-2y-3z|+|x-2y|+|x-3z| with x>0, y>0, z>0,

In[2]:=

Plot3D[Abs[4x-y-3]+Abs[x-2y]+Abs[4x+3y-3],{x,0,1},{y,0,1}]

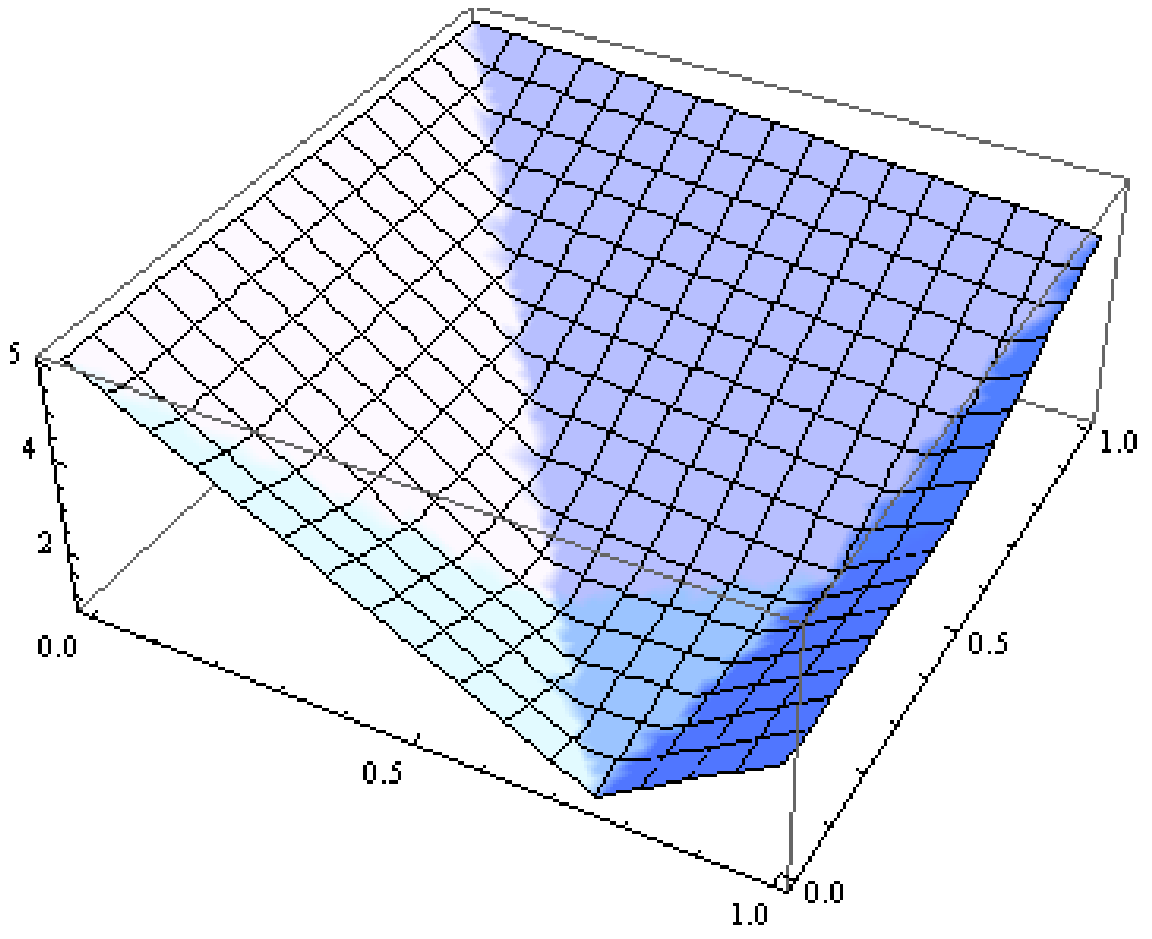

Then find the minimum of g(x,y) if any:

In[3]:=

FindMinValue[{Abs[4x-y-3]+Abs[x-2y]+Abs[4x+3y-3],x+y≤1,x>0,y>0},{x,y}]

The following result is returned:

Out[3]:= 0.841235.

FindMinValue::eit: The algorithm does not converge to the tolerance of 4.806217383937354`*^-6 in 500 iterations. The best estimated solution, with feasibility residual, KKT residual, or complementary residual of {0.0799888, 0.137702,0.0270028}, is returned.

## Matrix Method of using α-Discounting

The determinant of the homogeneous linear system (1), (2), (3) is:

$$\begin{vmatrix} 1 & -2\alpha_1 & -3\alpha_2 \\ -\frac{1}{2}\alpha_3 & 1 & 0 \\ -\frac{1}{3}\alpha_4 & 0 & 1 \end{vmatrix} = (1+0+0)-(\alpha_2\alpha_4+\alpha_1\alpha_3)=0$$

or

$$\alpha_1\alpha_3+\alpha_2\alpha_4=1 \text{ (parametric equation).}$$

The determinant has to be zero in order for the system to have non-null solutions.

The rank of the matrix is 2.

So, we find two variables, for example it is easier to solve for $y$ and $z$ from the last two equations, in terms of $x$:

$$\begin{cases} y = \frac{1}{2}\alpha_3 x \\ z = \frac{1}{3}\alpha_4 x \end{cases}$$

and the procedure follows the same steps as in the previous one.

Let us change Example 1 in order to study various situations.

## Weak Inconsistent Example 3

This is more weakly inconsistent than Example 2.

1. Same as in Example 2.
2. $C2$ is 4 times as important as $C1$
3. Same as in Example 2.

$$\begin{cases} x = 2\alpha_1 y + 3\alpha_2 z \\ y = 4\alpha_3 x \\ z = \frac{\alpha_4}{3} x \end{cases}$$

$$x = 2\alpha_1 (4\alpha_3 x) + 3\alpha_2 \left(\frac{\alpha_4}{3}\right) x$$

$$1 \cdot x = (8\alpha_1\alpha_3 + \alpha_2\alpha_4) \cdot x$$

$$8\alpha_1\alpha_3 + \alpha_2\alpha_4 = 1 \text{ (parametric equation)}$$

$$\alpha_1 = \alpha_2 = \alpha_3 = \alpha_4 = \alpha > 0.$$

$$9\alpha^2 = 1 \Rightarrow \alpha = \frac{1}{3}$$

$$\begin{bmatrix} x & 4\alpha_3 x & \frac{\alpha_4}{3} x \end{bmatrix} \rightarrow \begin{bmatrix} 1 & 4\alpha_3 & \frac{\alpha_4}{3} \end{bmatrix}$$

$$\begin{bmatrix} 1 & \frac{4}{3} & \frac{1}{9} \end{bmatrix} = \begin{bmatrix} \frac{9}{9} & \frac{12}{9} & \frac{1}{9} \end{bmatrix};$$

normalized: $\begin{bmatrix} \frac{9}{22} & \frac{12}{22} & \frac{1}{22} \end{bmatrix}$.

$$\frac{y}{x} = 1.333 \text{ instead of } 4;$$

$$\frac{z}{x} = 0.111 \text{ instead of } 0.3333;$$

$$x = 0.667y + 1 \cdot z \text{ instead of } 2y + 3z.$$

Each coefficient was reduced at $\frac{1}{3}(=33.33\%)$.

The bigger is the inconsistency $(\beta \to 1)$, the bigger is the discounting $(\alpha \to 0)$.

## Weak Inconsistent Example 4

This is even more inconsistent than Example 3.

1. Same as in Example 2.
2. Same as in Example 3.
3. $C3$ is 5 times as important as $C1$.

$$\begin{cases} x = 2\alpha_1 y + 3\alpha_2 z \\ y = 4\alpha_3 x \\ z = 5\alpha_4 x \end{cases}$$

$$x = 2\alpha_1(4\alpha_3 x) + 3\alpha_2(5\alpha_4 x)$$

$$1 \cdot x = (8\alpha_1\alpha_3 + 15\alpha_2\alpha_4)x$$

whence $8\alpha_1\alpha_3 + 15\alpha_2\alpha_4 = 1$ (parametric equation).

$$\alpha_1 = \alpha_2 = \alpha_3 = \alpha_4 = \alpha > 0,\ \ 23\alpha^2 = 1,\ \alpha = \frac{\sqrt{23}}{23}$$

$$\begin{bmatrix}1 & 4\alpha_3 & 5\alpha_4\end{bmatrix} \rightarrow \begin{bmatrix}1 & \frac{4\sqrt{23}}{23} & \frac{5\sqrt{23}}{23}\end{bmatrix}$$

Normalized: $\begin{bmatrix}0.34763 & 0.28994 & 0.36243\end{bmatrix}$

$\frac{y}{x} \cong 0.83405$ instead of 4, i.e. reduced at $\frac{\sqrt{23}}{23} = 20.85\%$

$$\frac{z}{x} \cong 1.04257 \text{ instead of } 5$$

$$x \cong 0.41703y + 0.62554 \cdot z \text{ instead of } 2x + 3y.$$

Each coefficient was reduced at $\alpha = \frac{\sqrt{23}}{23} \cong 20.85\%$.

## Consistent Example 5

When we get $\alpha = 1$, we have a consistent problem.

Suppose the preferences:

1. Same as in Example 2.
2. $C2$ is one fourth as important as $C1$.
3. $C3$ is one sixth as important as $C1$.

The system is:

$$\begin{cases} x = 2y + 3z \\ y = \frac{x}{4} \\ z = \frac{x}{6} \end{cases}$$

## First Method of Solving this System

Replacing the second and third equations of this system into the first, we get:

$$x = 2\left(\frac{x}{4}\right) + 3\left(\frac{x}{6}\right) = \frac{x}{2} + \frac{x}{2} = x,$$

which is an identity (so, no contradiction).

General solution:

$$\begin{bmatrix} x & \frac{x}{4} & \frac{x}{6} \end{bmatrix}$$

Priority vector:

$$\begin{bmatrix} 1 & \frac{1}{4} & \frac{1}{6} \end{bmatrix}$$

Normalized is:

$$\begin{bmatrix} \frac{12}{17} & \frac{3}{17} & \frac{2}{17} \end{bmatrix}$$

## Second Method of Solving this System

Let us parameterize:

$$\begin{cases} x = 2\alpha_1 y + 3\alpha_2 z \\ y = \frac{\alpha_3}{4} x \\ z = \frac{\alpha_4}{6} x \end{cases}$$

Replacing the last two equations into the first we get:

$$x = 2\alpha_1\left(\frac{\alpha_3}{4} x\right) + 3\alpha_2\left(\frac{\alpha_4}{6} x\right) = \frac{\alpha_1\alpha_3}{2} x + \frac{\alpha_2\alpha_4}{2} x$$

$$1 \cdot x = \frac{\alpha_1\alpha_3 + \alpha_2\alpha_4}{2} \cdot x,$$

whence $1 = \frac{\alpha_1\alpha_3 + \alpha_2\alpha_4}{2}$ or $\alpha_1\alpha_3 + \alpha_2\alpha_4 = 2$.

Consider the fairness principle: $\alpha_1 = \alpha_2 = \alpha_3 = \alpha_4 = \alpha > 0$, then $2\alpha^2 = 2$, $\alpha = \pm 1$, but we take only the positive value $\alpha = 1$ (as expected for a consistent problem).

Let us check:

$$\frac{y}{x} = \frac{\frac{3}{17}}{\frac{12}{17}} = \frac{1}{4}$$, exactly as in the original system;

$$\frac{z}{x} = \frac{\frac{2}{17}}{\frac{12}{17}} = \frac{1}{6}$$, exactly as in the original system;

$$x = 2y + 3z \text{ since } x = 2\left(\frac{x}{4}\right) + 3\left(\frac{x}{6}\right);$$

hence all coefficients were left at $\alpha = 1$ (=100%) of the original ones. No discount was needed.

## General Example 6

Let us consider the general case:

$$\begin{cases} x = a_1 y + a_2 z \\ y = a_3 x \\ z = a_4 x \end{cases}$$

where $a_1, a_2, a_3, a_4 > 0$

Let us parameterize:

$$\begin{cases} x = a_1\alpha_1 y + a_2\alpha_2 z \\ y = a_3\alpha_3 x \\ z = a_4\alpha_4 x \end{cases}$$

with $\alpha_1, \alpha_2, \alpha_3, \alpha_4 > 0$.

Replacing the second and third equations into the first, we get:

$$x = a_1\alpha_1\left(a_3\alpha_3 x\right) + a_2\alpha_2\left(a_4\alpha_4 x\right)$$

$$x = a_1a_3\alpha_1\alpha_3 x + a_2a_4\alpha_2\alpha_4 x$$

whence

$$a_1a_3\alpha_1\alpha_3 + a_2a_4\alpha_2\alpha_4 = 1 \text{ (parametric equation)}$$

The general solution of the system is:

$$\left(x,\ a_3\alpha_3 x,\ a_4\alpha_4 x\right)$$

The priority vector is $\left[1\ \ a_3\alpha_3\ \ a_4\alpha_4\right]$.

Consider the fairness principle:

$$\alpha_1 = \alpha_2 = \alpha_3 = \alpha_4 = \alpha > 0$$

We get:

$$\alpha^2 = \frac{1}{a_1a_3 + a_2a_4},$$

so,

$$\alpha = \frac{1}{\sqrt{a_1a_3 + a_2a_4}}$$

i) If $\alpha \in [0,1]$, then $\alpha$ is the degree of consistency of the problem, while $\beta = 1 - \alpha$ is the degree of the inconsistency of the problem.

ii) If $\alpha > 1$, then $\frac{1}{\alpha}$ is the degree of consistency, while $\beta = 1 - \frac{1}{\alpha}$ is the degree of inconsistency.

When the degree of consistency $\to 0$, the degree of inconsistency $\to 1$, and reciprocally.

## Discussion of the General Example 6

Suppose the coefficients $a_1, a_2, a_3, a_4$ become big such that $a_1a_3 + a_2a_4 \to \infty$, then $\alpha \to 0$, and $\beta \to 1$.

*Particular Example 7.*

Let us see a particular case when $a_1, a_2, a_3, a_4$ make $a_1a_3 + a_2a_4$ big:

$$a_1 = 50, \quad a_2 = 20, \quad a_3 = 100, \quad a_4 = 250,$$

then $\alpha = \dfrac{1}{\sqrt{50 \cdot 100 + 20 \cdot 250}} = \dfrac{1}{\sqrt{10000}} = \dfrac{1}{100} = 0.01$ = degree of consistency, whence $\beta = 0.99$ degree of inconsistency.

The priority vector for Particular Example 7 is:

$$\begin{bmatrix} 1 & 100(0.01) & 250(0.01) \end{bmatrix} = \begin{bmatrix} 1 & 1 & 2.5 \end{bmatrix}$$

which normalized is:

$$\begin{bmatrix} \frac{2}{9} & \frac{2}{9} & \frac{5}{9} \end{bmatrix}.$$

*Particular Example 8.*

Another case when $a_1, a_2, a_3, a_4$ make the expression $a_1a_3 + a_2a_4$ a tiny positive number:

$$a_1 = 0.02, \quad a_2 = 0.05, \quad a_3 = 0.03, \quad a_4 = 0.02, \text{ then}$$

$$\alpha = \frac{1}{\sqrt{0.02 \cdot (0.03) + 0.05 \cdot (0.02)}} = \frac{1}{0.04} = 25 > 1.$$

Then $\dfrac{1}{\alpha} = \dfrac{1}{25} = 0.04$ is the degree of consistency of the problem, and 0.96 the degree of inconsistency.

The priority vector for Particular Example 8 is:

$$\begin{bmatrix} 1 & a_3\alpha & a_4\alpha \end{bmatrix} = \begin{bmatrix} 1 & 0.03(25) & 0.02(25) \end{bmatrix} = \begin{bmatrix} 1 & 0.75 & 0.50 \end{bmatrix}$$

which normalized is $\begin{bmatrix} \frac{4}{9} & \frac{3}{9} & \frac{3}{9} \end{bmatrix}$.

Let us verify:

$\frac{y}{x} = \frac{3}{9} \div \frac{4}{9} = 0.75$ instead of 0.03, i.e. $\alpha = 25$ times larger (or 2500%);

$\frac{z}{x} = \frac{2}{9} \div \frac{4}{9} = 0.50$ instead of 0.02, i.e. 25 times larger;

$x = 0.50y + 1.25z$ instead of $x = 0.02y + 0.05z$ (0.50 is 25 times larger than 0.02, and 1.25 is 25 times larger than 0.05) because $\frac{4}{9} = 0.50\left(\frac{3}{9}\right) + 1.25\left(\frac{2}{9}\right)$.

## Jean Dezert's Weak Inconsistent Example 9

Let $\alpha_1, \alpha_2, \alpha_3 > 0$ be the parameters. Then:

$$\begin{cases} (5) \quad \frac{y}{x} = 3\alpha_1 \\ (6) \quad \frac{x}{z} = 4\alpha_2 \\ (7) \quad \frac{y}{z} = 5\alpha_3 \end{cases} \quad (5),(6) \Rightarrow \frac{y}{x} \cdot \frac{x}{z} = (3\alpha_1) \cdot (4\alpha_2) \Rightarrow \frac{y}{z} = 12\alpha_1\alpha_2$$

In order for $\frac{y}{z} = 12\alpha_1\alpha_2$ to be consistent with $\frac{y}{z} = 5\alpha_3$ we need to have $12\alpha_1\alpha_2 = 5\alpha_3$

$$\text{or } 2.4\alpha_1\alpha_2 = \alpha_3 \quad \text{(Parametric Equation)} \qquad (8)$$

Solving this system:

$$\begin{cases} \dfrac{y}{x} = 3\alpha_1 \Rightarrow y = 3\alpha_1 \cdot x \\ \dfrac{x}{z} = 4\alpha_2 \Rightarrow x = 4\alpha_2 \cdot z \\ \dfrac{y}{z} = 5\alpha_3 \Rightarrow y = 12\alpha_1\alpha_2 z \end{cases}$$

we get the general solution:

$$\begin{bmatrix} 4\alpha_2 z & 5(2.4\alpha_1\alpha_2)z & z \end{bmatrix}$$

$$\begin{bmatrix} 4\alpha_2 z & 12\alpha_1\alpha_2 z & z \end{bmatrix}$$

General normalized priority vector is:

$$\begin{bmatrix} \dfrac{4\alpha_2}{4\alpha_2 + 12\alpha_1\alpha_2 + 1} & \dfrac{12\alpha_1\alpha_2}{4\alpha_2 + 12\alpha_1\alpha_2 + 1} & \dfrac{1}{4\alpha_2 + 12\alpha_1\alpha_2 + 1} \end{bmatrix}$$

where $\alpha_1, \alpha_2 > 0$; ($\alpha_3 = 2.4\alpha_1\alpha_2$).

Which $\alpha_1$ and $\alpha_2$ give the best result? How to measure it? This is the greatest challenge!

$\alpha$ -Discounting Method includes all solutions (all possible priority vectors which make the matrix consistent).

Because we have to be consistent with all proportions (i.e. using the Fairness Principle of finding the parameters' numerical values), there should be the same discounting of all three proportions (5), (6), and (7), whence

$$\alpha_1 = \alpha_2 = \alpha_3 > 0 \qquad (9)$$

The parametric equation (8) becomes $2.4\alpha_1^2 = \alpha_1$ or

$$2.4\alpha_1^2 - \alpha_1 = 0,\ \alpha_1(2.4\alpha_1 - 1) = 0,$$

whence $\alpha_1 = 0$ or $\alpha_1 = \dfrac{1}{2.4} = \dfrac{5}{12}$.

$\alpha_1 = 0$ is not good, contradicting (9).

Our system becomes now:

$$\begin{cases} \frac{y}{x} = 3 \cdot \frac{5}{12} = \frac{15}{12} & (10) \\ \frac{x}{z} = 4 \cdot \frac{5}{12} = \frac{20}{12} & (11) \\ \frac{y}{z} = 5 \cdot \frac{5}{12} = \frac{25}{12} & (12) \end{cases}$$

We see that (10) and (11) together give

$$\frac{y}{x} \cdot \frac{x}{z} = \frac{15}{12} \cdot \frac{20}{12} \text{ or } \frac{y}{z} = \frac{25}{12},$$

so, they are now consistent with (12).

From (11) we get $x = \frac{20}{12} z$ and from (12) we get $y = \frac{25}{12} z$.

The priority vector is:

$$\begin{bmatrix} \frac{20}{12} z & \frac{25}{12} z & 1z \end{bmatrix}$$

which is normalized to:

$$\frac{\frac{20}{12}}{\frac{20}{12} + \frac{25}{12} + 1} = \frac{\frac{20}{12}}{\frac{20}{12} + \frac{25}{12} + \frac{12}{12}} = \frac{20}{57}, \quad \frac{\frac{25}{12}}{\frac{57}{12}} = \frac{25}{57}, \quad \frac{1}{\frac{57}{12}} = \frac{12}{57}, \text{ i.e.}$$

$$\begin{matrix} C_1 & C_2 & C_3 \end{matrix}$$

$$\begin{bmatrix} \frac{20}{57} & \frac{25}{57} & \frac{12}{57} \end{bmatrix}^T \qquad (13)$$

$$\begin{matrix} C_1 & C_2 & C_3 \end{matrix}$$

$$\cong \begin{bmatrix} 0.3509 & 0.4386 & 0.2105 \end{bmatrix}^T$$

↑ (0.4386 circled) the highest priority

Let us study the result:

$$\begin{matrix} C_1 & C_2 & C_3 \end{matrix}$$
$$\begin{bmatrix} \frac{20}{57} & \frac{25}{57} & \frac{12}{57} \end{bmatrix}^T$$
$$\begin{matrix} x & y & z \end{matrix}$$

*Ratios:* *Percentage of Discounting:*

$$\frac{y}{x} = \frac{\frac{25}{57}}{\frac{20}{57}} = \frac{25}{20} = 1.25 \text{ instead of } 3; \quad \frac{\frac{25}{20}}{3} = \frac{5}{12} = \alpha_1 = 41.\overline{6}\%$$

$$\frac{x}{z} = \frac{\frac{20}{57}}{\frac{12}{57}} = \frac{20}{12} = \frac{5}{3} = 1.\overline{6} \text{ instead of } 4; \quad \frac{\frac{20}{12}}{4} = \frac{5}{12} = \alpha_1 = 41.\overline{6}\%$$

$$\frac{y}{z} = \frac{\frac{25}{57}}{\frac{12}{57}} = \frac{25}{12} = 2.08\overline{3} \text{ instead of } 5; \quad \frac{\frac{25}{12}}{5} = \frac{5}{12} = \alpha_1 = 41.\overline{6}\%$$

Hence all original proportions, which were respectively equal to 3, 4, and 5 in the problem, were reduced by multiplication with the same factor $\alpha_1 = \frac{5}{12}$, i.e. by getting $41.\overline{6}\%$ of each of them.

So, it was fair to reduce each factor to the same percentage $41.\overline{6}\%$ of itself.

But this is not the case in Saaty's method: its normalized priority vector is

$$\begin{matrix} C_1 & C_2 & C_3 \\ [0.2797 & 0.6267 & 0.0936]^T, \\ x & y & z \end{matrix}$$

where:

*Ratios:* *Percentage of Discounting:*

$\frac{y}{x} = \frac{0.6267}{02797} \cong 2.2406$ instead of 3; $\frac{2.2406}{3} \cong 74.6867\%$

$\frac{x}{z} = \frac{0.2797}{0.0936} \cong 2.9882$ instead of 4; $\frac{2.9882}{4} \cong 74.7050\%$

$\frac{y}{z} = \frac{06267}{0.0936} \cong 6.6955$ instead of 5; $\frac{6.6955}{5} \cong 133.9100\%$

Why, for example, the first proportion, which was equal to 3, was discounted to $74.6867\%$ of it, while the second proportion, which was equal to 4, was discounted to another percentage (although close) $74.7050\%$ of it?

Even more doubt we have for the third proportion's coefficient, which was equal to 5, but was increased to $133.9100\%$ of it, while the previous two proportions were decreased; what is the justification for these?

That is why we think our α-D/Fairness-Principle is better justified.

We can solve this same problem using matrices. (5), (6), (7) can be written in another way to form a linear parameterized homogeneous system:

$$\begin{cases} 3\alpha_1 x - y = 0 \\ x - 4\alpha_2 z = 0 \\ y - 5\alpha_3 z = 0 \end{cases} \quad (14)$$

whose associated matrix is:

$$P_1 = \begin{bmatrix} 3\alpha_1 & -1 & 0 \\ 1 & 0 & -4\alpha_2 \\ 0 & 1 & -5\alpha_3 \end{bmatrix} \qquad (15)$$

a) If $\det(P_1) \neq 0$ then the system (10) has only the null solution $x = y = z = 0$.

b) Therefore, we need to have $\det(P_1) = 0$, or $(3\alpha_1)(4\alpha_2) - 5\alpha_3 = 0$, or $2.4\alpha_1\alpha_2 - \alpha_3 = 0$, so we get the same parametric equation as (8).

In this case the homogeneous parameterized linear system (14) has a triple infinity of solutions.

This method is an extension of Saaty's method, since we have the possibility to manipulate the parameters $\alpha_1, \alpha_2$, and $\alpha_3$. For example, if a second source tells us that $\frac{x}{z}$ has to be discounted 2 times as much as $\frac{y}{x}$, and $\frac{y}{z}$ should be discounted 3 times less than $\frac{y}{x}$, then we set $\alpha_2 = 2\alpha_1$, and respectively $\alpha_3 = \frac{\alpha_1}{3}$, and the original (5), (6), (7) system becomes:

$$\begin{cases} \frac{y}{x} = 3\alpha_1 \\ \frac{x}{z} = 4\alpha_2 = 4(2\alpha_1) = 8\alpha_1 \\ \frac{y}{z} = 5\alpha_3 = 5\left(\frac{\alpha_1}{3}\right) = \frac{5}{3}\alpha_1 \end{cases} \qquad (16)$$

and we solve it in the same way.

## Weak Inconsistent Example 10

Let us complicate Jean Dezert's Weak Inconsistent Example 9. with one more preference: $C_2$ is 1.5 times as much as $C_1$ and $C_3$ together.

The new system is:

$$\begin{cases} \frac{y}{x} = 3 \\ \frac{x}{z} = 4 \\ \frac{y}{z} = 5 \\ y = 1.5(x+z) \\ x, y, z \in [0,1] \\ x + y + z = 1 \end{cases} \qquad (17)$$

We parameterized it:

$$\begin{cases} \frac{y}{x} = 3\alpha_1 \\ \frac{x}{z} = 4\alpha_2 \\ \frac{y}{z} = 5\alpha_3 \\ y = 1.5\alpha_4(x+z) \\ x, y, z \in [0,1] \\ x + y + z = 1 \end{cases} \qquad (18)$$

$$\alpha_1, \alpha_2, \alpha_3, \alpha_4 > 0$$

Its associated matrix is:

$$P_2 = \begin{bmatrix} 3\alpha_1 & -1 & 0 \\ 1 & 0 & -4\alpha_2 \\ 0 & 1 & -5\alpha_3 \\ 1.5\alpha_4 & -1 & 1.5\alpha_4 \end{bmatrix} \qquad (19)$$

The rank of matrix $P_2$ should be strictly less than 3 in order for the system (18) to have non-null solution.

If we take the first three rows in (19) we get the matrix $P_1$ whose determinant should be zero, therefore one also gets the previous parametric equation $2.4\alpha_1\alpha_2 = \alpha_3$.

If we take rows 1, 3, and 4, since they all involve the relations between $C_2$ and the other criteria $C_1$ and $C_3$ we get

$$P_3 = \begin{bmatrix} 3\alpha_1 & -1 & 0 \\ 0 & 1 & -5\alpha_3 \\ 1.5\alpha_4 & -1 & 1.5\alpha_4 \end{bmatrix} \qquad (20)$$

whose determinant should also be zero:

$$\det(P_3) = \left[3\alpha_1 \ (1.5\alpha_4) + 5\alpha_3(1.5\alpha_4) + 0\right] - \left[0 + 3\alpha_1(5\alpha_3) + 0\right] =$$

$$= 4.5\alpha_1\alpha_4 + 7.5\alpha_3\alpha_4 - 15\alpha_1\alpha_3 = 0 \qquad (21)$$

If we take

$$P_4 = \begin{bmatrix} 1 & 0 & -4\alpha_2 \\ 0 & 1 & -5\alpha_3 \\ 1.5\alpha_4 & -1 & 1.5\alpha_4 \end{bmatrix} \qquad (22)$$

Then

$$\det(P_4) = [1.5\alpha_4 \ + 0 + 0] - [-6\alpha_2\alpha_4 + 5\alpha_3 + 0] = 1.5\alpha_4 + 6\alpha_2\alpha_4 - 5\alpha_3 = 0 \quad (23)$$

If we take

$$P_5 = \begin{bmatrix} 3\alpha_1 & -1 & 0 \\ 1 & 0 & -4\alpha_2 \\ 1.5\alpha_4 & -1 & 1.5\alpha_4 \end{bmatrix} \qquad (24)$$

then

$$\det(P_5) = [0 + 0 + 6\alpha_2\alpha_4] - [0 + 12\alpha_1\alpha_2 - 1.5\alpha_4] = 6\alpha_2\alpha_4 - 12\alpha_1\alpha_2 + 1.5\alpha_4 = 0 \quad (25)$$

So, these four parametric equations form a parametric system:

$$\begin{cases} 2.4\alpha_1\alpha_2 - \alpha_3 = 0 \\ 4.5\alpha_1\alpha_4 + 7.5\alpha_3\alpha_4 - 15\alpha_1\alpha_3 = 0 \\ 1.5\alpha_4 + 6\alpha_2\alpha_4 - 5\alpha_3 = 0 \\ 6\alpha_2\alpha_4 - 12\alpha_1\alpha_2 + 1.5\alpha_4 = 0 \end{cases} \qquad (26)$$

which should have a non-null solution.

If we consider $\alpha_1 = \alpha_2 = \alpha_3 = \frac{5}{12} > 0$ as we got at the beginning, then substituting all α's into the last three equations of the system (26) we get:

$$4.5\left(\frac{5}{12}\right)\alpha_4 + 7.5\left(\frac{5}{12}\right)\alpha_4 - 15\left(\frac{5}{12}\right)\left(\frac{5}{12}\right) = 0 \Rightarrow \alpha_4 = 0.5208\bar{3} = \frac{25}{48}$$

$$1.5\alpha_4 + 6\left(\frac{5}{12}\right)\alpha_4 - 5\left(\frac{5}{12}\right) = 0 \Rightarrow \alpha_4 = 0.5208\bar{3}$$

$$6\left(\frac{5}{12}\right)\alpha_4 - 12\left(\frac{5}{12}\right)\left(\frac{5}{12}\right) + 1.5\alpha_4 = 0 \Rightarrow \alpha_4 = 0.5208\bar{3}$$

$\alpha_4$ could not be equal to $\alpha_1 = \alpha_2 = \alpha_3$ since it is an extra preference, because the number of rows was bigger than the number of columns. So the system is consistent, having the same solution as previously, without having added the fourth preference $y = 1.5(x + z)$.

## Jean Dezert's Strong Inconsistent Example 11

The preference matrix is:

$$M_1 = \begin{pmatrix} 1 & 9 & \frac{1}{9} \\ \frac{1}{9} & 1 & 9 \\ 9 & \frac{1}{9} & 1 \end{pmatrix}$$

so,

$$\begin{cases} x = 9y, x > y \\ x = \frac{1}{9}z, x < z \\ y = 9z, y > z \end{cases}$$

The other three equations: $y = \frac{1}{9}x$, $z = 9x$, $z = \frac{1}{9}y$ result directly from the previous three ones, so we can eliminate them.

From x>y and y>z (first and third above inequalities) we get x>z, but the second inequality tells us the opposite: x<z; that is why we have a strong contradiction/inconsistency. Or, if we combine all three we have x>y>z>x… strong contradiction again.

Parameterize:

$$\begin{cases} x = 9\alpha_1 y & (27) \\ x = \frac{1}{9}\alpha_2 z & (28) \\ y = 9\alpha_3 z & (29) \end{cases}$$

where $\alpha_1, \alpha_2, \alpha_3 > 0$.

From (27) we get: $y = \frac{1}{9\alpha_1}x$, from (28) we get $z = \frac{1}{9\alpha_2}x$, which is replaced in (29) and we get:

$$y = 9\alpha_3\left(\frac{9}{\alpha_2}x\right) = \frac{81\alpha_3}{\alpha_2}x.$$

So $\frac{1}{9\alpha_1}x = \frac{81\alpha_3}{\alpha_2}x$ or $\alpha_2 = 729\alpha_1\alpha_3$ (parametric equation).

The general solution of the system is:

$$\left(x,\ \frac{1}{9\alpha_1}x,\ \frac{9}{\alpha_2}x\right)$$

The general priority vector is:

$$\left[1\ \ \frac{1}{9\alpha_1}\ \ \frac{9}{\alpha_2}\right].$$

Consider the fairness principle, then $\alpha_1 = \alpha_2 = \alpha_3 = \alpha > 1$ are replaced into the parametric equation: $\alpha = 729\alpha^2$, whence $\alpha = 0$ (not good) and $\alpha = \frac{1}{729} = \frac{1}{9^3}$.

The particular priority vector becomes

$$\left[1\quad 9^2\quad 9^4\right] = \left[1\quad 81\quad 6561\right]$$

and normalized

$$\left[\frac{1}{6643}\quad \frac{81}{6643}\quad \frac{6561}{6643}\right]$$

Because the consistency is $\alpha = \frac{1}{729} = 0.00137$ is extremely low, we can disregard this solution (and the inconsistency is very big $\beta = 1 - \alpha = 0.99863$).

## Remarks

a) If in $M_1$ we replace all six 9's by a bigger number, the inconsistency of the system will increase. Let us use 11.

Then $\alpha = \frac{1}{11^3} = 0.00075$ (consistency), while inconsistency $\beta = 0.99925$.

b) But if in $M_1$ we replace all 9's by the smaller positive number greater than 1, the consistency decreases. Let us use 2. Then $\alpha = \frac{1}{2^3} = 0.125$ and $\beta = 0.875$;

c) Consistency is 1 when replacing all six 9's by 1.

d) Then, replacing all six 9's by a positive sub unitary number, consistency decreases again. For example, replacing by 0.8 we get $\alpha = \frac{1}{0.8^3} = 1.953125 > 1$, whence $\frac{1}{\alpha} = 0.512$ (consistency) and $\beta = 0.488$ (inconsistency).

## Jean Dezert's Strong Inconsistent Example 12

The preference matrix is:

$$M_2 = \begin{pmatrix} 1 & 5 & \frac{1}{5} \\ \frac{1}{5} & 1 & 5 \\ 5 & \frac{1}{5} & 1 \end{pmatrix}$$

which is similar to $M_1$ where we replace all six 9's by 5's.

$\alpha = \frac{1}{5^3} = 0.008$ (consistency) and $\beta = 0.992$ (inconsistency).

The priority vector is $\begin{bmatrix} 1 & 5^2 & 5^4 \end{bmatrix} = \begin{bmatrix} 1 & 25 & 625 \end{bmatrix}$

and normalized $\begin{bmatrix} \frac{1}{651} & \frac{25}{651} & \frac{625}{651} \end{bmatrix}$.

$M_2$ is a little more consistent than $M_1$ because 0.00800 > 0.00137, but still not enough, so this result is also discarded.

## Generalization of Jean Dezert's Strong Inconsistent Examples

*General Example 13.*

Let the preference matrix be:

$$M_t = \begin{pmatrix} 1 & t & \frac{1}{t} \\ \frac{1}{t} & 1 & \text{t} \\ \text{t} & \frac{1}{t} & 1 \end{pmatrix},$$

with $t > 0$, and $c(M_t)$ the consistency of $M_t$, $i(M_t)$ inconsistency of $M_t$.

We have for the Fairness Principle:

$\lim_{t\to 1} c(M_t) = 1$ and $\lim_{t\to 1} i(M_t) = 0$;

$\lim_{t\to +\infty} c(M_t) = 0$ and $\lim_{t\to +\infty} i(M_t) = 1$;

$\lim_{t\to 0} c(M_t) = 0$ and $\lim_{t\to 0} i(M_t) = 1$.

Also $\alpha = \frac{1}{t^3}$, the priority vector is $\begin{bmatrix} 1 & t^2 & t^4 \end{bmatrix}$ which is normalized as

$$\begin{bmatrix} \frac{1}{1+t^2+t^4} & \frac{t^2}{1+t^2+t^4} & \frac{t^4}{1+t^2+t^4} \end{bmatrix}.$$

In such situations, when we get strong contradiction of the form x>y>z>x or similarly x<z<x, etc. and the consistency is tiny, we can consider that x=y=z=1/3 (so no criterion is preferable to the other – as in Saaty's AHP), or just x+y+z=1 (it means that one has the total ignorance too: $C1 \cup C2 \cup C3$).

## Strong Inconsistent Example 14

Let C = {C1, C2}, and P = {C1 is important twice as much as C2; C2 is important 5 times as much as C1}. Let m(C1)=x, m(C2)=y. Then: x=2y and y=5x (it is a strong inconsistency since from the first equation we have x>y, while from the second y>x).

Parameterize: $x=2\alpha_1 y$, $y=5\alpha_2 x$, whence we get $2\alpha_1=1/(5\alpha_2)$, or $10\alpha_1\alpha_2=1$.

If we consider the Fairness Principle, then $\alpha_1= \alpha_2= \alpha>0$, and one gets $\alpha = \frac{\sqrt{10}}{10} \approx 31.62\%$ consistency; priority vector is [0.39 0.61], hence y>x. An explanation can be done as in paraconsistent logic (or as in neutrosophic logic): we consider that the preferences were honest, but subjective, therefore it is possible to have two contradictory statements true simultaneously since a criterion C1 can be more important from a point of view than C2, while from another point of view C2 can be more important than C1. In our decision-making problem, not having any more information and having rapidly being required to take a decision, we can prefer C2, since C2 is 5 times more important that C1, while C1 is only 2 times more important than C2, and 5>2.

If it's no hurry, more prudent would be in such dilemma to search for more information on C1 and C2.

If we change Example 14 under the form: x=2y and y=2x (the two coefficients set equal), we get α = ½, so the priority vector is [0.5 0.5] and decision-making problem is undecidable.

## Non-Linear Equation System Example 15

Let C = {C1, C2, C3}, m(C1)=x, m(C2)=y, m(C3)=z.

Let F be:

1. C1 is twice as much important as the product of C2 and C3.
2. C2 is five times as much important as C3.

We form the non-linear system: x=2yz (non-linear equation) and y=5z (linear equation).

The general solution vector of this mixed system is: $[10z^2 \quad 5z \quad z]$, where z>0.

A discussion is necessary now.

a) You see for sure that y>z, since 5z>z for z strictly positive. But we don't see anything what the position of x would be?

b) Let us simplify the general solution vector by dividing each vector component by z>0, thus we get: $[10z \quad 5 \quad 1]$.

If $z \in (0, 0.1)$, then y>z>x.

If z=0.1, then y>z=x.

If $z \in (0.1, 0.5)$, then y>x>z.

If z=0.5, then y=x>z.

If z>0.5, then x>y>z.

## Non-Linear/Linear Equation/Inequality Mixed System Example 16

Since in the previous Example 15 has many variants, assume that a new preference comes in (in addition to the previous two preferences):

1. C1 is less important than C3.

The mixed system becomes now: $x=2yz$ (non-linear equation), $y=5z$ (linear equation), and $x<z$ (linear inequality).

The general solution vector of this mixed system is: $[10z^2 \;\; 5z \;\; z]$, where $z>0$ and $10z^2 < z$. From the last two inequalities we get $z \in (0, 0.1)$. Whence the priorities are: $y>z>x$.

## Future Research

To investigate the connection between α-D MCDM and other methods, such as: the technique for order preference by similarity to ideal solution (TOPSIS) method, the simple additive weighting (SAW) method, Borda-Kendall (BK) method for aggregating ordinal preferences, and the cross-efficiency evaluation method in data envelopment analysis (DEA).

## Conclusion

We have introduced a new method in the multi-criteria decision making, $\alpha$ - Discounting MCDM. In the first part of this method, each preference is transformed into a linear or non-linear equation or inequality, and all together form a system that is resolved – one finds its general solution, from which one extracts the positive solutions. If the system has only the null solution, or it is inconsistent, then one parameterizes the coefficients of the system.

In the second part of the method, one chooses a principle for finding the numerical values of the parameters {we have proposed herein the Fairness Principle, or Expert's Opinion on Discounting, or setting a Consistency (or Inconsistency) Threshold}.

## Acknowledgement

The author wants to thank Mr. Atilla Karaman, from the Operations Research Department of KHO SAVBEN in Ankara, Turkey, Ph D student using this α-Discounting Method for Multi-Criteria Decision Making in his Ph D thesis, for his remarks on this chapter.

# Three Non-linear α-Discounting MCDM-Method Examples

## Abstract

In this chapter we present three new examples of using the α-Discounting Multi-Criteria Decision Making Method in solving non-linear problems involving algebraic equations and inequalities in the decision making process.

## Introduction

We have defined a new procedure called *α-Discounting Method for Multi-Criteria Decision Making (α-D MCDM)*, which is as an alternative and extension of Saaty's Analytical Hierarchy Process (AHP). We have also defined the *degree of consistency* (and implicitly a *degree of inconsistency*) of a decision-making problem [1].

The α-D MCDM can deal with any set of preferences that can be transformed into an algebraic system of linear and/or non-linear homogeneous and/or non-homogeneous equations and/or inequalities.

We discuss below three new examples of non-linear decision making problems.

## Example 1

The Set of References is $C_1, C_2, C_3$ .

1. $C_1$ is as important as the product of $C_2$ and $C_3$.
2. The square of $C_2$ is as important as $C_3$.
3. $C_3$ is less important than $C_2$.

We denote $C_1 = x,\ C_2 = y,\ C_3 = z$, and we'll obtain the following non-linear algebraic system of two equations and one inequality:

$$\begin{cases} x = yz \\ y^2 = z \\ z < y \end{cases}$$

and of course the conditions

$$\begin{cases} x + y + z = 1 \\ x, y, z \in \ 0,1 \end{cases}$$

From the first two equations we have:

$$x = yz = y \cdot y^2 = y^3$$

and $z = y^2$ whence the general priority vector is $\langle y^3\ y\ \ y^2 \rangle$.

We consider $y \neq 0$, because if $y = 0$ the priority vector becomes $\langle 0\ 0\ 0 \rangle$ which does not make sense.

Dividing by $y$ we have $\langle y^2\ 1\ y \rangle$, and normalized:

$$\left\langle \frac{y^2}{y^2+y+1}\quad \frac{1}{y^2+y+1}\quad \frac{y}{y^2+y+1} \right\rangle.$$

From $y^2 = z$ and $z < y$ we have

$y^2 < y$ or $y^2 - y < 0$ or $y\ \ y-1\ < 0$, hence $y \in\ 0,1$ .

For $y \in \;\; 0,1$ we have the order:

$$\frac{1}{y^2+y+1} > \frac{y}{y^2+y+1} > \frac{y^2}{y^2+y+1}$$

so $C_2 > C_3 > C_1$.

## Example 2

The Set of References is also $C_1, C_2, C_3$ .

1. $C_1$ is as important as the square of $C_2$.
2. $C_2$ is as important as double $C_3$.
3. The square of $C_3$ is as important as triple $C_1$.

We again denote $C_1 = x$, $C_2 = y$, $C_3 = z$, and we'll obtain the following non-linear algebraic system of three equations:

$$\begin{cases} x = y^2 \\ y = 2z \\ z^2 = 3x \end{cases}$$

If we solve it, we get:

$$\left.\begin{matrix} x = 4z^2 \\ x = \dfrac{z^2}{3} \end{matrix}\right\} \Rightarrow 4z^2 = \frac{z^2}{3} \Rightarrow z = 0 \Rightarrow y = 0 \Rightarrow x = 0.$$

Algebraically the only solution is $\langle 0\ 0\ 0 \rangle$, but the null solution is not convenient for MCDM.

Let's parameterize, i.e. "discount" each equality:

$$\left.\begin{cases} \left.\begin{matrix} x = \alpha_1 y^2 \\ y = 2\alpha_2 z \end{matrix}\right\} \Rightarrow x = \alpha_1 \;\; 2\alpha_2 z \;\;^2 = 4\alpha_1\alpha_2^2 z^2 \\ z^2 = 3\alpha_3 x \Rightarrow x = \dfrac{1}{3\alpha_3} z^2 \end{cases}\right\} \Rightarrow 4\alpha_1\alpha_2^2 = \frac{1}{3\alpha_3}$$

or $12\alpha_1\alpha_2^2\alpha_3 = 1$, which is the characteristic parametric equation needed for the consistency of this algebraic system.

The general algebraic solution in this parameterized case is: $<4\alpha_1\alpha_2^2 z^2 \quad 2\alpha_2 z \quad z>$.

Using the Fairness Principle as in the α-Discounting MCDM Method, we set all parameters equal:

$$\alpha_1 = \alpha_2 = \alpha_3 = \alpha$$

whence, from the characteristic parametric equation, we obtain that

$$12\alpha^4 = 1,$$

therefore

$$\alpha = \sqrt[4]{\frac{1}{12}} = \frac{1}{\sqrt[4]{12}} \approx 0.537.$$

Thus the general solution for the Fairness Principle is:

$$<4\alpha^3 z^2 \quad 2\alpha z \quad z>$$

and, after substituting $\alpha \approx 0.537$, it results:

$$\begin{cases} x = 4\alpha^3 z^2 \simeq 0.619 z^2 \\ y = 2\alpha z = 1.074 z. \end{cases}$$

Whence the general solution for the Fairness Principle becomes:

$$\langle 4\alpha^3 z^2 \quad 2\alpha z \quad z \rangle \simeq \langle 0.619 z^2 \quad 1.074 z \quad z \rangle$$

and dividing by $z \neq 0$ one has:

$$\langle 0.619z,\ 1.074,\ 1 \rangle.$$

But y = 1.074 > 1 = z, hence y > z.

*Discussion:*

1. If $z < \frac{1}{0.619} \approx 1.616$, then y > z > x.

2. If $z = \frac{1}{0.619} \approx 1.616$, then y > z = x.
3. If $\frac{1}{0.619} < z < \frac{1.074}{0.619}$ or 1.616 < z < 1.735, then y > x > z.
4. If $z = \frac{1.074}{0.619} \approx 1.735$, then x = y > z.
5. If $z > \frac{1.074}{0.619} \approx 1.735$, then x > y > z.

From the orders of x, y, and z it results the corresponding orders between the preferences $C_1$, $C_2$, and $C_3$.

## Example 3

Let's suppose that the sources are not equally reliable.

First source is five times less reliable than the second, while the third source is twice more reliable that the second one. Then the parameterized system:

$$\begin{cases} x = \alpha_1 y^2 \\ y = 2\alpha_2 z \\ z^2 = 3\alpha_3 x \end{cases} \text{ becomes } \begin{cases} x = 3\alpha_2 y^2 \\ y = 2\alpha_2 z \\ z^2 = 3\frac{\alpha_2}{4} x \end{cases}$$

since $\alpha_1 = 3\alpha_2$ which means that we need to discount the first equation three times more than the second, and $\alpha_3 = \frac{\alpha_2}{4}$ which means that we need to discount the third equation a quarter of the second equation's discount.

Denote $\alpha_2 = \alpha$, then:

$$\begin{cases} x = 3\alpha y^2 \\ y = 2\alpha z \\ z^2 = \dfrac{3\alpha}{4} x \end{cases}$$

whence

$$x = 3\alpha(2\alpha z)^2 = 12\alpha^3 z^2$$

and

$$x = \frac{4}{3\alpha} z^2$$

therefore

$$12\alpha^3 z^2 = \frac{4}{3\alpha} z^2, \text{ or } 36\alpha^4 = 4$$

Thus

$$\alpha = \frac{1}{\sqrt[4]{9}} \simeq 0.485.$$

The algebraic general solution is:

$$\langle 12\alpha^3 z^2 \quad 2\alpha z \quad z \rangle = \langle 1.557z^2 \quad 0.970z \quad z \rangle = \left\langle \frac{1.557z}{1.557z + 1.854} \quad 0.854 \quad 1 \right\rangle.$$

And in a similar way, as we did for Example 2, we may discuss upon parameter $z > 0$ the order of preferences $C_1$, $C_2$, and $C_3$.

# Interval α-Discounting Method for MDCM

## Abstract

This chapter is an extension of our previous work on α-Discounting Method for MCDM ([1], [2], [3]) from crisp numbers to intervals.

## Introduction

In 2010 we have introduced a new method [3], called α-Discounting Method for Multi Criteria Decision Making, which is an alternative but also a generalization of Saaty's Analytical Hierarchy Process (AHP). α-Discounting Method works for any $n$-pairwise comparisons, $n \geq 2$, that may be linear or non-linear, or may be equations or inequalities. It transforms all preferences into a system of equations and/or of inequalities, that is later solved algebraically.

Since *Saaty's AHP* is not the topic of this chapter, we'll not present over here. Neither our *α-Discounting Method for Multi Criteria Decision Making* is recalled. The interested reader may get them in [7], and respectively [1], [2], and [3].

## A consistent example

Let have the set of criteria be $C = \{C_1, C_2, C_3\}$, and the set of preferences $P$ be:

1. $C_1$ is twice or three times as important as $C_2$;
2. $C_2$ is one or one and half times as important as $C_3$.

### Solution

Let $x$ represents the value of $C$, $y$ of $C_2$, and $z$ of $C_3$. $x > 0, y > 0, z > 0$.

We form the algebraic interval system:

$$\begin{cases} x = [2,3]y \\ y = [1,1.5]z, \end{cases}$$

where $[2,3]$ and $[1,1.5]$ are intervals.

Replacing the second equation into the first, one gets:

$$x = [2,3]y = [2,3] \cdot [1,1.5]z = [2 \cdot 1, 3 \cdot 1.5]z = [2,4.5]z.$$

The general solution of this system is:

$$\langle [2,4.5]z, [1,1.5]z, z \rangle \text{ where } z > 0.$$

We divide this vector components by $z$, and we get:

$$\underset{C_1}{\langle [2,4.5]}, \underset{C_2}{[1,1.5]}, \underset{C_3}{1\rangle}.$$

We don't know exactly what to mean by normalization when dealing with intervals, but it is clear that $C_1 > C_2 > C_3$.

## A second consistent example

Criteria: $C = \{C_1, C_2, C_3\}$, and the set of preferences $P$, same as in the previous example, but adding one more:

1. $C_1$ is twice or three times as important as $C_2$;
2. $C_2$ is one or one and half times as important as $C_3$;
3. $C_3$ is $\frac{1}{4}$ or $\frac{1}{2}$ times as important as $C_1$ .

## Solution

With same notations $x, z$ and $z$ representing the values of $C_1, C_2$ and respectively $C_3$, we form the algebraic system:

$$\begin{cases} x = [2,3] \cdot y \\ y = [1,1.5] \cdot z \\ z = [0.25, 0.50] \cdot x, \end{cases}$$

with $x > 0, y > 0, z > 0$.

The determinant of the system is:

$$\begin{vmatrix} 1 & -[2,3] & 0 \\ 0 & 1 & -[1,1.5] \\ -[0.25, 0.50] & 0 & 1 \end{vmatrix} = \begin{vmatrix} 1 & [-3,-2] & 0 \\ 0 & 1 & [-1{,}5,-1] \\ [-0.50,-0.25] & 0 & 1 \end{vmatrix}$$

$$= 1 — [-3,-2] \cdot [-0.50, -0.25]$$
$$= [1,1] + ][(-3) \cdot (-1.59) \cdot (-0.5), (2) \cdot (-1) \cdot (-0.25)] = [1,1] + [-2.25, -0.50]$$
$$= [1 - 2.25, 1 - 0.50] = [-1.25, 0.50]$$
$$\neq [0,0].$$

Let's parameterize the system, using $\alpha_1 > 0, \alpha_2 > 0, \alpha_3 > 0$ in order to discount each interval coefficient. We get:

$$\begin{cases} x = \alpha_1 [2,3] y \\ y = \alpha_2 [1,1.5] z \\ z = \alpha_3 [0.25, 0.50] x. \end{cases}$$

The determinant of the parameterized system is:

$$\begin{vmatrix} 1 & -\alpha_1 [2,3] & 0 \\ 0 & 1 & -\alpha_2 [1,1.5] \\ -\alpha_3 [0.25, 0.50] & 0 & 1 \end{vmatrix}$$

$$= 1 - \alpha_1 \alpha_2 \alpha_3 [2,3][1,1.5][0.25, 0.50]$$
$$= [1,1] - \alpha_1 \alpha_2 \alpha_3 [0.50, 2.25] = [0{,}0].$$

Whence $\alpha_1 \alpha_2 \alpha_3 [0.50, 2.25] = [1,1]$, hence $\alpha_1 \alpha_2 \alpha_3 = \left[\frac{1}{2.25}, \frac{1}{0.50}\right] = \left[\frac{4}{9}, 2\right]$.

For equitable discount, let $\alpha_1 = \alpha_2 = \alpha_3 = \alpha > 0$. Then

$$\alpha^3 = \left[\frac{4}{9}, 2\right], \text{whence } \alpha = \left[\sqrt[3]{4/9}, \sqrt[3]{2}\right] \simeq [0{,}76, 1.26].$$

There, the system is altered with the same proportion $\alpha = [0{,}76, 1.26]$ each equation, and it becomes:

$$\begin{cases} x = [2,3] \cdot \alpha \cdot y = [2,3] \cdot [0.76, 1.26]y = [1.52, 3.78]y \\ y = [1, 1.5] \cdot \alpha \cdot z = [1, 1.5] \cdot [0.76, 1.26]z = [0.76, 1.89]z \\ z = [0.25, 0.50] \cdot \alpha \cdot x = [0.25, 0.50] \cdot [0.76, 1.26]x = [0.19, 0.63]x \end{cases}$$

From the first two equations we get:

$$\begin{cases} x = [1.52, 3.78]y = [1.52, 3.78] \cdot [0.76, 1.89]z = [1.16, 7.14]z \\ y = [0.76, 1.89]z \end{cases}$$

The third equation

$$1 \cdot z = [0.19, 0.63]x$$

is equivalent to

$$x = \frac{1}{[0.19, 0.63]}z = \left[\frac{1}{0.63}, \frac{1}{0.19}\right]z = [1.59, 5.26]z.$$

Therefore we got the following approximation that we can call reconciliation of the first equations together, that give us:

$$x = [1.16, 7.14]z$$

with respect to the third equation that gives us:

$$x = [1.59, 5.26]z.$$

We see that the intervals $[1.16, 7.14]$ and $[1.59, 5.26]$ are close to each other.

The solution vector of the parameterized system, for $\alpha = [0.76, 1.26]$ is

$$\langle [1.16, 7.14]z \text{ or} [1.59, 5.26]z, [0.76, 1.89]z, z \rangle.$$

We divide by $z > 0$ and we get:

$$\langle \underset{C_1}{[1.16, 7.14] \text{ or } [1.59, 5.26]}, \underset{C_2}{[0.76, 1.89]}, \underset{C_3}{1} \rangle.$$

It's not necessary to normalize. We can see that:

$$C_1 > C_2 \text{ and } C_1 > C_{3.}$$

To compare $C_2$ with $C_3$, we see that in general

$$C_2 > C_3,$$

since the interval $[0.76, 1.89]$ has a bigger part which is

$$(1, 1.89] > 1 \text{ when } C_2 > C_3,$$

and a smaller part $[0.76, 1)$ when $C_2 > C_3$, while a single case $[1, 1] = 1$ when $C_2 = C_3$.

## Inconsistent example

Same criteria and the first two preferences. Only the third preference is changed as in the below third equation.

$$\begin{cases} x = [2, 3]y \\ y = [1, 1.5]z \\ z = [3, 3.5]x. \end{cases}$$

From first and second equations, we get

$$x = [2, 3]y = [2, 3], [1, 1.5]z = [2, 4.5]z.$$

From the last equation:

$$x = \frac{1}{[3, 3.5]} = \left[\frac{1}{3.5}, \frac{1}{3}\right] \simeq [0.29, 0.33]z$$

which is different from $[2, 4.5]z$.

Parameterized in the same way as before:

$$\begin{cases} x = \alpha_1[2, 3]y \\ y = \alpha_2[1, 1.5]z \\ z = \alpha_3[3, 3.5]x. \end{cases}$$

We similarly get from the first two equations:

$$x = c[2, 4.5]z$$

and from the last equation:

$$x = \frac{1}{[3, 3.5]} \cdot \frac{1}{\alpha_3} \simeq \frac{1}{\alpha_3}[0.29, 0.33]z.$$

Whence:

$$\alpha_1\alpha_2[2, 4.5]z = \frac{1}{\alpha_3}[0.29, 0.33]z$$

$$\text{Or } \alpha_1\alpha_2\alpha_3[2, 4.5] = [0.29, 0.33],$$

hence

$$\alpha_1\alpha_2\alpha_3 = \frac{[0.29, 0.33]}{2, 4.5]} = \left[\frac{0.29}{4.5}, \frac{0.33}{2}\right] \simeq [0.064, 0.165].$$

Considering an equitable discount we set

$$\alpha_1 = \alpha_2 = \alpha_3 = \alpha > 0,$$

hence $\alpha_1\alpha_2\alpha_3 = [0.064, 0.165]$

becomes $\alpha^3 = [0.064, 0.165]$,

whence $\alpha = \left[\sqrt[3]{0.064}, \sqrt[3]{0.165}\right] \simeq [0.400, 0.548]$.

Whence we get:

$$x = \alpha_1\alpha_2[2, 4.5]z = [0.400, 0.548][0.400, 0.548][2, 4.5]z$$
$$= [0.32, 1.55]z$$

or

$$x = \frac{1}{\alpha_3}[0.29, 0.33]z = \frac{1}{[0.400, 0.548]} \cdot [0.29, 0.33]z$$
$$= \left[\frac{0.24}{0.548}, \frac{0.33}{0.400}\right] z = [0.529, 0.825]z$$

and

$$y = \alpha_2[1, 1.5]z = [0.400, 0.548][1, 1.5]z = [0.400, 0.822]z.$$

The solution of the parameterized system is:

$$\langle [0.32, 1.55]z \text{ or } [0.529, 0.825]z, [0.400, 0.822]z, z\rangle.$$

We divide by $z > 0$ and we get:

$$\langle \underset{C_1}{[0.32, 1.55] \text{or } [0.529, 0.825]}, \underset{C_2}{[0.400, 0.822]}, \underset{C_3}{1}\rangle.$$

Clearly $C_2 < C_3$.

Then $C_1 < C_3$ for most part of its values, i.e. for $[0.32, 1)$ or $[0.529, 0.825]$ and $C_1 > C_3$for (1, 1.55) while $C_1 = C_3$ for [1, 1]. To compare $C_1$ and $C_2$ it is more complicated.

## Conclusion

In this chapter we have constructed two consistent examples and one inconsistent example of decision making problems, where the preferences use intervals instead of crisp numbers in comparisons of preferences. The results are, of course, more complicated.

In this book we introduce a new procedure called α-Discounting Method for Multi-Criteria Decision Making (α-D MCDM), which is as an alternative and extension of Saaty's Analytical Hierarchy Process (AHP). It works for any number of preferences that can be transformed into a system of homogeneous linear equations. A degree of consistency (and implicitly a degree of inconsistency) of a decision-making problem are defined. α-D MCDM is afterwards generalized to a set of preferences that can be transformed into a system of linear and/or non-linear homogeneous and/or non-homogeneous equations and/or inequalities.

The general idea of α-D MCDM is to assign non-null positive parameters $\alpha_1, \alpha_2, \ldots, \alpha_p$ to the coefficients in the right-hand side of each preference that diminish or increase them in order to transform the above linear homogeneous system of equations which has only the null-solution, into a system having a particular non-null solution. After finding the general solution of this system, the principles used to assign particular values to all parameters α's is the second important part of α-D, yet to be deeper investigated in the future.

In the current book we propose the Fairness Principle, i.e. each coefficient should be discounted with the same percentage (we think this is fair: not making any favoritism or unfairness to any coefficient), but the reader can propose other principles.

For consistent decision-making problems with pairwise comparisons, α-Discounting Method together with the Fairness Principle give the same result as AHP.

But for weak inconsistent decision-making problem, α-Discounting together with the Fairness Principle give a different result from AHP.

Many consistent, weak inconsistent, and strong inconsistent examples are given in this book.